\documentclass[a4paper,conference]{IEEEtran}
\pdfoutput=1
\usepackage{cite}
\usepackage{algorithmic}
\usepackage{algorithm}
\usepackage{graphicx}
\usepackage{subfigure}
\usepackage{float}
\usepackage{threeparttable}
\usepackage{multirow}
\usepackage{multicol}
\usepackage{float}
\usepackage{xcolor}
\usepackage{amsmath}

\definecolor{Yellow}{rgb}{1,0.9,0.7}
\definecolor{Pink}{rgb}{1,0.85,0.85}
\definecolor{AntiqueWhite}{rgb}{0.9,0.9,0.9}

\newcommand{\NOTE}[1]%
{
\noindent
\fboxsep=2mm\fcolorbox{black}{AntiqueWhite}{\parbox{0.95\columnwidth}
{\textbf{NOTE: } #1}
}
}
\begin{document}

\title{Dynamic Resource-aware Corner Detection for Bio-inspired Vision Sensors}

\author{\IEEEauthorblockN{Sherif A.S. Mohamed\IEEEauthorrefmark{1},
Jawad N. Yasin\IEEEauthorrefmark{1},
Mohammad-hashem Haghbayan\IEEEauthorrefmark{1},
Antonio Miele \IEEEauthorrefmark{2},
Jukka Heikkonen\IEEEauthorrefmark{1}, \\
Hannu Tenhunen\IEEEauthorrefmark{3},
and Juha Plosila\IEEEauthorrefmark{1}}
\IEEEauthorblockA{\IEEEauthorrefmark{1}Department of Future Technologies, University of Turku, 20500 Turku, Finland}
\IEEEauthorblockA{\IEEEauthorrefmark{2}Department of Industrial and Medical Electronics, Royal Institute of Technology (KTH), 16440 Kista, Sweden}
\IEEEauthorblockA{\IEEEauthorrefmark{3}Dipartimento di Elettronica, Informazione e Bioingegneria, Politecnico di Milano, 20133 Milano, Italy}}

\maketitle

\begin{abstract}
Event-based cameras are vision devices that transmit only brightness changes with low latency and ultra-low power consumption. Such characteristics make event-based cameras attractive in the field of localization and object tracking in resource-constrained systems. Since the number of generated events in such cameras is huge, the selection and filtering of the incoming events are beneficial from both increasing the accuracy of the features and reducing the computational load. In this paper, we present an algorithm to detect asynchronous corners form a stream of events in real-time on embedded systems. The algorithm is called the Three Layer Filtering-Harris or TLF-Harris algorithm. The algorithm is based on an events' filtering strategy whose purpose is 1) to increase the accuracy by deliberately eliminating some incoming events, i.e., noise and 2) to improve the real-time performance of the system, i.e., preserving a constant throughput in terms of input events per second, by discarding unnecessary events with a limited accuracy loss.
An approximation of the Harris algorithm, in turn, is used to exploit its high-quality detection capability with a low-complexity implementation to enable seamless real-time performance on embedded computing platforms. The proposed algorithm is capable of selecting the best corner candidate among neighbors and achieves an average execution time savings of $59\%$ compared with the conventional Harris score. Moreover, our approach outperforms the competing methods, such as eFAST, eHarris, and FA-Harris, in terms of real-time performance, and surpasses Arc* in terms of accuracy. 
\end{abstract}

\IEEEpeerreviewmaketitle

\section{Introduction}

Different types of computer vision-based methods, such as Visual Simultaneous Localization and  Mapping (vSLAM), Optical Flow (OF) estimation, obstacle avoidance, and object tracking rely heavily on the quality of the extracted corners and edges in images. Edges and corners are points that have a unique position and can be easily detected in a sequence of images \cite{sherif, jdj1, jdj2}. Corners are one of the most unique features in the image since they show strong changes in all eight directions which makes them different and easy to differentiate from the neighboring points. For example, most of the vSLAM approaches in the state-of-the-art \cite{vslam1,vslam2,vslam3,vslam4} are based on analyzing images captured by traditional frame-based cameras to obtain the translation and rotation between two consecutive images. Traditional cameras or frame-based cameras use CMOS sensors to generate grayscale or RGB images at a fixed rate (e.g. 60 frames per second, or FPS) by capturing the absolute intensity of all pixels simultaneously. Extracting consistent and robust corners is very challenging from frame-based cameras since they are affected by motion blur in highly dynamic scenes and they have blind spots, i.e., the time between two consecutive frames.  Moreover, images captured by conventional cameras contain a lot of redundant information that increases the computational complexity dramatically.  Thus,   vision tasks based on such cameras are unlikely to achieve real-time performance on low-cost embedded systems due to the limited resources.

Newly emerged bio-inspired vision sensors, i.e., event-based cameras, offer high potential to overcome the challenging task of detecting high frequency and consistent corners in complex scenarios, such as highly dynamic and low-illuminance environments \cite{sherif}. Event cameras trigger events by capturing only brightness changes with a high temporal resolution, i.e., they can trigger 10 million events per second in some scenarios. Moreover, they have low power consumption in terms of 10 mW and they can operate in low lighting conditions since they have high dynamic range, i.e., 140 dB \cite{eventsurvey}. These attractive characteristics make event cameras an ideal solution for applications that require to perform real-time vision-based algorithms on resource-constrained systems, i.e., embedded systems.

\begin{figure}
    \centering
    \includegraphics[width=.9\columnwidth]{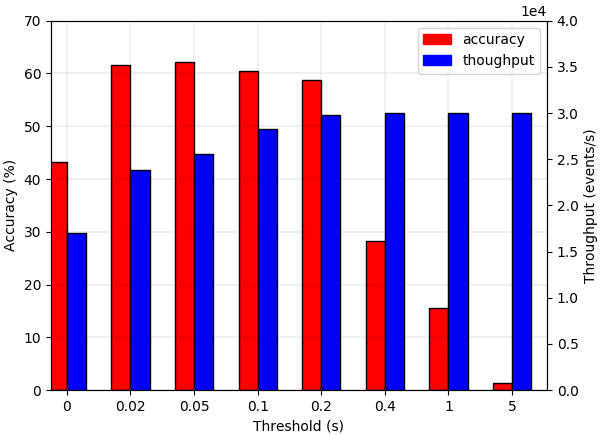}
    \caption{Relation between the accuracy and the throughput vs. the timestamp threshold. }
    \label{fig:acc}
\end{figure}{}

Two factors determine the quality in the processing of an approach based event cameras; the \textit{accuracy} in the detection of the corners, and the \textit{throughput}, that we here measure in terms of the number of events the system is able to process per unit of time.
Figure~\ref{fig:acc} shows the accuracy and throughput vs. the timestamp threshold used to filter the event (0s = we keep all the events, while 5s = the event is discarded if another previous event occurred in the same position less than 5 seconds before.).
On one hand, it can be noticed that both the low filtering rate of the events and over-filtering negatively affects the accuracy. 
On the other hand, the number of maintained events, i.e. not filtered away, determines the computation load and performance of the processing system; indeed if this number of events is too high, the system may be not able to achieve the desired throughput requirement, also considering that other applications may concurrently run on the architectural platform and therefore compete for the processing resources. As a conclusion, we may notice how the events' filtering may be a suitable knob for co-optimizing both accuracy and throughput of systems based on event cameras.

We present, in this paper, a novel corner detection algorithm which consists of three layers of filtering. The filtering process considers both the type of the events, to improve the accuracy, and the load they cause on the system, to preserve the desired throughput, at the same time. More precisely, our contributions in this paper is a novel \textbf{three-layer-filtering}, able to efficiently maintain a real-time performance by reducing the event-stream and selecting only the most meaningful and unique events to be processed. The selection technique is based on the characteristics of the incoming events and its relation to the neighboring events in the local patch and the status of the hardware parts (e.g. CPU), by monitoring the throughput of the embedded system. Our method is able to run 11x faster than eHarris and 2x faster than FA-Harris. Moreover, our method performs favorably compared to other methods in the state-of-the-art in terms of accuracy.
    
In Section 2, we highlight relevant works in the literature. Afterword, we describe the proposed algorithm in Section 3. In Section 4, we summarize the experimental setup and results. Lastly, we draw our conclusion. 

\section{Related Work}

Corner detection is a core part of various computer vision algorithms, such as pose and depth estimation, object tracking, and obstacle avoidance. It is used to extract points of interest, i.e., corners from a sequence of images. The reason for using corners in many vision-based algorithms is that they are invariant to brightness changes and motions changes, i.e, orientation and translation. These characteristics make it possible to robustly detect and easily distinguish corners points from neighboring points under various circumstances. Corner detection methods can be categorized into two main types: 1) methods that use frame-based cameras and 2) methods based on  event cameras.

\subsection{Intensity-based corner detection}

These methods use the intensity value of each pixel in images to detect corners. These images are captured using a traditional camera, i.e., a frame-based camera which uses a CMOS sensor to capture the absolute intensity of each pixel. A simple method to detect corners in an intensity image is by using correlation. However, this method is suboptimal and computationally expensive. There are several methods that have been presented in the past years to extract strong corners from images. For instance, Harris \cite{harris} and SIFT \cite{sift} use difference of Gaussian (DoG) to detect corners, SURF \cite{surf}, and FAST \cite{fast} uses a Bresenham circle of radius 3 around each point, i.e., a circle of 16 pixels to extract corners with low computational complicity. Generally, such methods present various drawbacks since they are based on images captured by frame-based cameras to detect corners. Such cameras provide blurry images in high motion speed scenarios, which negatively affects the quality of detected corners. Moreover, the traditional camera transmits the absolute intensity for all pixels in the sensor to generate images, and thus images might contain a lot of redundant information which increases the computational complexity dramatically.

Event-based methods use the characteristic of events to extract corners from pre-generated event-frames, i.e., indirect methods or directly from asynchronous events, i.e., direct methods. 

\noindent\textbf{Indirect methods:} They extract corners from event-frames using traditional frame-based corner detectors, for example, Harris \cite{harris}. Event-frames are generated by gathering events in a fixed or dynamic fashion.  In the fixed fashion, event-frames can be generated in two different ways: 1) by gathering events during a specific time period \cite{timeWindow} or 2) by gathering a certain number of events \cite{numberEvent}.  Another way is to dynamically set the temporal window required to generate a frame. For instance, in \cite{lifeTime} the temporal window is based on the lifetime of the event. In \cite{sherif2}, authors presented an algorithm to set the temporal window dynamically based on the amount of the information in the scene, i.e., entropy and the camera motion. In general, such methods present various drawbacks based on the complexity of the environment and the motion speed, such as generating blurry event-frames due to over-accumulating events or generating noisy frames by under-accumulating the sufficient amount of events to reconstruct the scene. Moreover, generating event-frames omit one of the inherent characteristics of event cameras, i.e., events are triggered in an asynchronous manner. 

\begin{figure*}[t]
    \centering
    \vspace{5pt}
    \includegraphics[width=\textwidth]{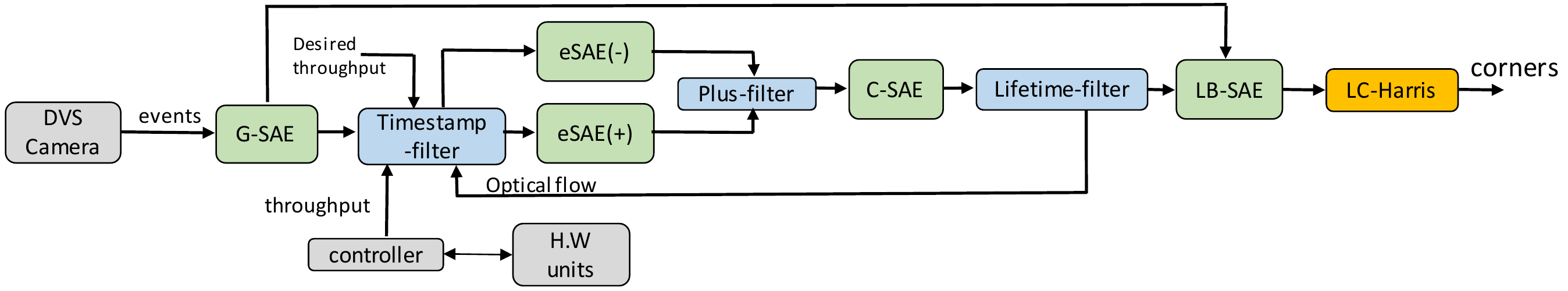}
    \caption{System overview of the proposed TLF-Harris detector. The three-layer filtering is represented in blue and the SAE data representations are in green. }
    \label{fig:system}
\end{figure*}

\noindent\textbf{Direct methods:} 
They extract corners directly from a stream of asynchronous events to exploit all characteristics of event cameras. In \cite{eharris}, Vasco et al. proposed an algorithm to detect asynchronous corners using an adaptation of the original Harris score \cite{harris} on Surface Active Events (SAE). The algorithm shows high-quality corner detection, however, it demands a lot of computational resources to compute the gradients of each incoming event. Scheerlinck et al. \cite{scheerlinck} used a high-pass filter and a 3x3 convolution matrix, i.e., kernel to reconstruct the scene asynchronously upon the arrival of each incoming event. To detect corners the authors used two Sobel kernels in $S_x$ and $S_y$ to calculate the gradient in x and y-axis respectively. Afterward, the Harris score is used to classify if the incoming event is an actual corner.

In \cite{eFAST}, the authors present an algorithm inspired by FAST \cite{fast} to detect event-corners, also referred to as eFAST. In this method, the event-corners are extracted by comparing the timestamp of the latest events, i.e., pixels on the inner and the outer circle which has a radius of 3 and 4 pixels respectively. The algorithm searches for an \textit{arc} of length equal to 3-6 pixels on the inner circle and an \textit{arc} of the length of 4-8 pixels on the outer circle to classify an incoming event as an actual corner, i.e., event-corner. Arc is a segment of continuous pixels that the most recent events from the rest pixels on the circle. In \cite{arc}, the authors used a similar technique (Arc) with an event filter to detect corner corners 4.5x faster than the eFAST. Both methods are shown to run faster than the Harris-based method since the corner detection is based on pixel-wise comparisons and avoiding expensive operations. However, the accuracy of the detection of such methods is poor compared with methods that use Harris.
In conclusion, extracting events directly from events is essential to exploit all attributes of event cameras. However, there is a trade-off between the real-time performance and the accuracy of corner detection. Since event cameras can generate millions of events in a second and processing each incoming event would increase the computational complexity dramatically. On the other hand, methods that rely on simple pixel-wise comparisons to achieve real-time performance produce poor corners. Therefore, we present our algorithm which consists of three-layer filtering to achieve high accuracy and real-time performance on embedded systems.

\section{Proposed Approach}

The proposed Three Layer Filtering-Harris algorithm, shortly TLF-Harris, for corner detection, is depicted in Fig.~\ref{fig:system}.
Into the system, events are represented by means of the Surface Active Events (SAE) model\cite{eFAST}. Each event taken from the camera is processed by means of a filtering unit composed of three steps, i.e., dynamic Timestamp-filter, Plus-filter, and Lifetime-filter. The three-layer filtering reduces the number of redundant events to improve real-time performance and increase the accuracy by discarding outliers. The filtered events are transmitted to a low-complexity \textit{corner selector} unit, called $LC-Harris$, which computes the score of the incoming event to classify it as a corner or not a corner. The pipeline of the four phases is executed in sequence continuously on the series of input events, in which arrival frequency may vary during the time. Then, the throughput is computed as the number of input events the system is able to process per unit of time. 

In the following subsections, we highlight the model of the event camera and the SAE representation. Then, we will provide a comprehensive description of each unit in the proposed algorithm.

\subsection{Event-Based Camera Model}
DVS camera \cite{dvs} is a bio-inspired, lightweight, and fast vision sensor that has interesting attributes, such as low latency, high output rate, HDR, robust to motion blur, and low power consumption, making them ideal for embedded systems and fast-moving micro-robotics. The event camera generates a stream of events in an asynchronous manner when each pixel's brightness changes over a certain value, i.e., brightness threshold. Each event contains four simple information and to extract useful information from these events they are typically converted to time-surface representation. An event $e = (u,v,pol,ts)$ contains the 2D position $(u,v)$ of the pixel, the polarity $pol$ of the event showing the increase ($+1$) or decrease ($-1$) in pixel brightness, and the timestamp $ts$ of the event. An event occurs when the brightness $I$ of any pixel of the sensor, measured in two subsequent time instants with a small sampling resolution $\triangle{t}$, increases or decreases in a logarithmic scale by the amount of the brightness threshold $C$, typically adjusted between 15\%$\sim$50\%. Formally, the event is generated when the following inequality becomes true:
\begin{equation}
    \log(I((u,v),ts)) - log(I((u,v), ts-\triangle{t})) \geq pol*C
\end{equation}

\subsection{Surface Active Events}
The elaboration on the incoming events is mainly performed on the relations between each event with the neighbor ones and on their occurrence times.
To do this one common way is to use the Spatio-temporal domain in a Surface Active Events (SAE) representation \cite{sae2}. The SAE can be considered as an elevation map, which composed of three values. The location of the event $e(u,v)$ in the image and the timestamp $ts$. The SAE is updated asynchronously when an event is triggered

\begin{figure}[t]
\centering
\includegraphics[width= .8\columnwidth]{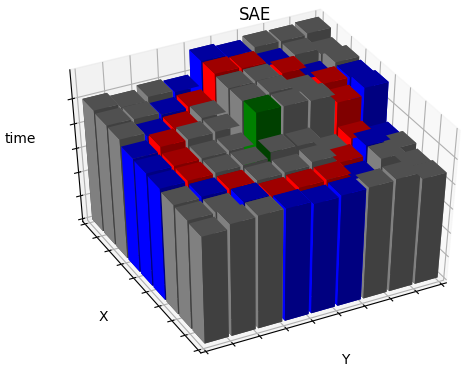}
\caption{The local Surface Active Events (SAE) of size 9x9, the inner (red) and outer (blue) circles of radius 3 and 4 pixels, respectively.}
\label{fig:sae}
\end{figure}

\begin{equation}
    SAE:(u_{i}, v_{i}) \rightarrow ts_{i}
\end{equation}
Figure~\ref{fig:sae} shows an example of SAE; in particular, for the sake of simplicity, it depicts a local SAE, i.e., capturing the information of a 9x9 sub-part of the image.

Within the proposed approach presented in Fig.~\ref{fig:system}, we use different types of SAE to perform the various filtering tasks and select strong corners based on the relation of the incoming event and neighboring events. In particular, we use four different types of SAEs: global, enhanced, corner, and local-binary SAEs. The global (G-SAE) has the same size as the DVS sensor, i.e., 240x180 and contains all incoming events. The enhanced SAE (eSAE) accumulates events based on the polarity of the incoming event: $eSAE(+)$ is used to accumulate events with positive polarities and $SAE(-)$ is for events with negative polarities. The corner SAE (C-SAE) is updated by the Plus-filter and contains only the timestamp of corner candidates. The local-binary SAE (LB-SAE) with a size of 9x9 is used to calculate the gradient of the incoming event and contains the most \textit{N} recent neighboring events labeled as 1's.

\subsection{First-Layer Filter: Timestamp Filter}

\begin{figure}[b]
\centering
    \includegraphics[width= .8\columnwidth]{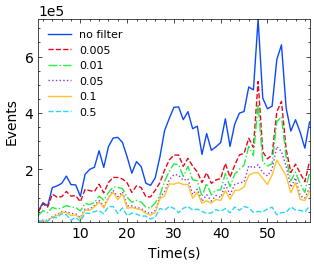}
    \caption{The effect of various timestamp thresholds on the event stream. Slow camera motion between second [0 - 30] and fast camera motion between second [30 - 60].}
    \label{fig:manyTime}
\end{figure}{}

The main purpose of the first filter (called \textit{timestamp-filter}) is to discard redundant events and maintain a balanced computational load across different environmental conditions, due to variations in the number of input events to be processed during the time.
Removing redundant events is not only beneficial for improving real-time performance but also enhance the quality of corner detection by removing outliers, see Figure~\ref{fig:acc}. For instance, a fast camera, object movement, or significant brightness changes would generate more than one event on the same pixel with different timestamps. Those triggered events do not provide any useful information, however, processing them would increase the computational load significantly as they come within a short period of time.

Inspired by the \cite{arc}, where the authors proposed a filter to discard redundant events from the processing pipeline. However, in this approach, the authors used a fixed threshold value (0.05 seconds). As illustrated in Figure \ref{fig:manyTime}, the main problem of using a fixed threshold is that a fixed threshold is not suitable for all camera movement speeds and environments. For instance, in fast camera a small threshold is recommended since using a large threshold would discard real events, i.e., inliers and thus degrade the extract performance. Note in \ref{fig:manyTime} at fast-motion the event stream curve is deformed by increasing the threshold. On the other hand, in slow motion, the curve still keeps a similar shape. Therefore, we propose a dynamic Timestamp-filter to dynamically adjust the timestamp threshold based on two parameters:  1) the movement speed of the camera and objects in the environment, i.e., optical flow and 2) the throughput of the system.

\begin{algorithm}[h]
\caption{dynamic Timestamp filter}
\small
\textbf{Input:} $G\_SAE$, $expected\_thr$, $OF_j$\\
\textbf{Output:} $eSAE(+), eSAE(-)$\\
\vspace{-.4cm}
\begin{algorithmic}[1]
\label{alg:timestamp}
\STATE $e_{old}.pol \leftarrow G\_SAE(e.u, e.v)$
\IF{$e.pol \neq e_{old}.pol$}
\STATE update $eSAE(e.pol)$
\ELSE
\STATE $OF_j \leftarrow batch(e.u, e.v)$
\IF{$e.OF > \theta$}
\STATE $ts_{new} = 0.01$
\ELSE
\STATE $thr_{app}\leftarrow get\_throughput()$\\
\IF{$thr_{app} < thr_{exp}$}
\STATE $ts_{new} = ts_{old} + K $
\ELSE
\STATE $ts_{new}= ts_{old} - K$
\ENDIF
\ENDIF
\STATE $e_{old}.ts \leftarrow G\_SAE(e.u, e.v)$
\IF{$e.ts > e_{old}.ts + ts_{new}$ }
\STATE update $eSAE(e.pol)$
\ENDIF
\ENDIF
\end{algorithmic}
\end{algorithm}

The first filtering phase of the proposed technique, called timestamp filter, is explained in Algorithm~\ref{alg:timestamp}. It continuously monitors the status of throughput and determines the \textit{timestamp threshold} based on the performance profiling (Lines 9-14). The timestamp threshold is used to update the $(eSAE(+), eSAE(-))$ pair by filtering the events captured from $G\_SAE$ (Lines 17-19). This process happens by comparing the polarity of the incoming event from $G\_SAE$ with the polarity of the previous event on the same pixel. If the polarity differs the eSAE is updated, otherwise the eSAE is only updated if the timestamp of the incoming event is larger than the summation of the previous event's timestamp and the filter threshold.  

The filter gets a stream of asynchronous events from the DVS camera, the required throughput, and monitors the status of the CPU units to generate an enhanced SAE $(eSAE(+), eSAE(-))$ (see Figure~\ref{fig:system}). The expected throughput can be set to the desired value. In case the application is under-performing, i.e., application throughput is less than the desired throughput, filter increases the timestamp threshold by a certain amount $K$ (e.g., 5ms) until the application achieve the desired throughput (Line 11). On the other hand, if the application is over-performing, i.e., the throughput is more than the desired throughput, the filter decreases the timestamp threshold (Line 13). To avoid discarding real events in high-speed cameras and objects moving in the scene, we divide the sensor into 108 batches of size 20x20. For each batch, we calculate the optical flow and obtain the optical flow of the incoming event by checking in which batch the event is located. If the event optical flow is more than a certain threshold $\theta$ the timestamp filter value is set to (0.01 s) regardless of the throughput of the system (Lines 5-7).

\subsection{Second-Layer Filter: Plus-Filter}

In order to reduce the number of events processed by the last stage, i..e, LC\_Harris, we implement a second layer filtering (called Plus-filter) filter inspired by the FAST \cite{fast}. As mentioned previously, the authors in \cite{eFAST} proposed an algorithm to extract corners from a stream of events by searching for a segment of continuous pixels of length [3-6] and [4-8] in two circles of radius 3 and 4 pixels respectively. The main issue of this approach is that it fails to detect corners with an angle greater than $180^\circ$. The authors in \cite{arc}, tackled this issue by presenting an algorithm that classifies an incoming event as a corner if the arc or its complementary arc on the two circles is between the threshold [3-6] for the small circle and [4-8] for the big circle. Both elements use several branching to obtain the length of the arc on the inner and the outer circles which increases the execution time. Therefore, we propose a light and fast spatial filter, i.e., Plus-filter to process only event-corner candidates to the next step in the pipeline. The filter extracts a local patch with size 7x7 around the incoming event from the eSAE. First, we extract the oldest and newest events on a circle of radius 3 pixels. Afterword, the 16 pixels on the circle are labeled clockwise starting from the newest event in ascending order. A set of four pixels $\gamma = {2, 6, 10, 14}$ is used to classify the incoming event is an actual event-corner candidate or not. If three out of the four pixels are bigger or smaller than the other pixel, the incoming event is considered as an event-corner candidate, i.e., processed to the next phase.   

\subsection{Third-Layer Filter: Lifetime Filter}

In order to tackle the effect of blurry event-corners (see Figure \ref{fig:observe}.a), or in other words, reduce the number of event-corners per intensity-corner. We implemented a third layer filtering based on the concept of event lifetime \cite{lifeTime}, thus called Lifetime filter. It is summarized in Algorithm~\ref{alg:lifetime}. The lifetime of an event is the time an event will take to shift by 1 pixel. The mechanism of the filter is to discard event-corner candidates that have timestamp lower than the lifetime of an existing event-corner in the local patch, i.e., neighborhood.  The filter search for a neighbor, i.e., event-corner within a Manhattan distance equal to 8 from the incoming event-corner candidate. If the incoming event-corner candidate has  timestamp higher than the lifetime of the neighboring event-corner, the lifetime is computed as

\begin{equation}
\label{eq:lifetime}
    \tau = max \bigtriangleup t \hspace{4pt} subject \hspace{4pt} to \parallel \bigtriangleup p \parallel= 1 \hspace{4pt} pixel
\end{equation}{}

where $\tau$ is the lifetime of the event, $p$ is the 2D position of the event, the displacement is denoted by $\parallel \bigtriangleup p \parallel$ and $\bigtriangleup t = SAE(p+\bigtriangleup p)-SAE(p)$.

Finally, the event goes to the last stage (LC\_Harris unit) to decide either it is a corner or not. Otherwise, it is removed from the pipeline.

\begin{algorithm}
\caption{Lifetime filter}
\small
\textbf{Input:} event-corner candidate $e=\{u,v,pol,ts\}$ \\
\vspace{-.4cm}
\begin{algorithmic}[1]
\label{alg:lifetime}
    \STATE Check Local Patch $CL\_SAE(e.u,e.v)$
    \IF{found neighbour}
        \STATE Check $life\_time$
        \IF{$ts > life\_time$}
            \STATE Delete old $life\_time$
            \STATE Calculate new $life\_time$
            \STATE Update $CL\_SAE$ 
            \STATE Process event
        \ENDIF
        \STATE discard event
    \ELSE
        \STATE Calculate $life\_time$
        \STATE Update $CL\_SAE$ 
        \STATE Process event
    \ENDIF
\end{algorithmic}
\end{algorithm}
    
\subsection{Low-complex Harris Score}  

Most computer vision approaches use Harris corner detector\cite{harris} to extract strong and consistent corners from traditional images. Since Harris detector is considered a high-performance corner detection operator that is commonly used in most computer vision algorithms. The Harris score $R$ of each pixel $p(u,v)$ is calculated based on the approximated matrix M of the local auto-correlation function: 

\begin{equation}
    M = \begin{bmatrix}
    \sum_{w}{G(p)I^{2}_{u}} & \sum_{w}{G(p)I_{u}I_{v}} \\ \sum_{w}{G(p)I_{u}I_{v}} & \sum_{w}{G(p)I^{2}_{v}}\end{bmatrix} = \begin{bmatrix} a & b \\ c & d\end{bmatrix}
\end{equation}{}

Where $I_{u}$ and $I_{v}$ are the horizontal and vertical gradients respectively. The local patch around the pixel $p(u,v)$ is represented by $w$ and $G(p)$ denotes a Gaussian filter. Harris used the eigenvalues $\lambda_{1}$ and $\lambda_{2}$ to compute the score as shown in (4), where $k$ is a constant and commonly set between 0.04 to 0.06. A point $p(u,v)$ is considered as a corner point if both $\lambda_{1}$ and $\lambda_{2}$ are large, in other words, the score $R$ is bigger than a certain threshold.

\begin{equation}
    R = \lambda_{1}\lambda_{2} - k*(\lambda_{1} + \lambda_{2})^2 = (ac - b^{2}) - k*(a+c)^{2}
\end{equation}{}

However, it is high performance demanding to compute the eigenvalues for each incoming event; as a consequence, such a technique is infeasible to run in real-time on embedded systems. This situation is exacerbated by the fact that event cameras can transmit more than 10 million events per second in high motion speeds scenarios. Thus, reducing the computational complexity is essential to be able to use the Harris detector for event camera on resource-constrained systems. 

Therefore, in this work, in addition to the three-layer of filtering to reduce the number of redundant events, we propose an approximation of the Harris algorithm. A binary local patch, that is a submatrix of the SAE around each incoming corner candidate with a size of 9x9 is extracted. Only the most recent $N$ neighbors (N = 25) are included and labeled as 1s in the local patch. The vertical and horizontal gradient is computed from the local binary batch and then used to calculate the Harris score:

\begin{equation}
   R = a^{'}*c^{'} = \sum|I_{x}|*\sum|I_{y}|
\end{equation}{}

\section{Experimental Evaluation}

\begin{table*}[ht]
\centering
\begin{threeparttable}
\caption{Comparison of different corner detection approaches on reduction percentage[\%].}\label{tab:filter}
\begin{tabular}{l|l|ccccc}
\hline                
                    
                      &\textbf{Dataset}
                      &\textbf{eFAST} \cite{eFAST} &\textbf{eHarris} \cite{eharris} &\textbf{Arc*} \cite{arc} &\textbf{FA-Harris} \cite{fa-harris}
                      &\textbf{Ours}\\ \hline
\multirow{2}{*}{low texture}
& shapes       & $87.00$  & $90.15$ & $88.33$  & $ 95.66$ & \textbf{99.06}\\    
&sun   &   $98.73$  & $95.51$ & $96.88$& $99.46$ & \textbf{99.63} \\  \hline
\multirow{2}{*}{high texture}
&boxes  &   $96.82$ & $92.25$  &$92.19$ & $97.95$  & \textbf{99.49} \\
&hdr\_poster   & $95.77$  & $93.09$ & $92.40$ &  $ 98.31$ & \textbf{99.55}\\\hline
\multirow{2}{*}{slow movement}
&outdoors\_walking  &  $97.91$ & $95.19$ & $95.53$&   $99.06$ & \textbf{99.61}\\

&office\_spiral & $97.04$ & $93.83$  &$93.53$ & $98.97$  & \textbf{99.09}\\\hline
\multirow{2}{*}{fast movement}
&night\_run         & $96.51$   &$94.50$ &$97.33$ &  $99.30$ &\textbf{99.84}\\

&bicycle  &  $96.08$  &$93.80$ & $96.53$ & $ 99.26$ & \textbf{99.75}\\ 
 \hline
\end{tabular}
\end{threeparttable}
\end{table*}

We evaluated the proposed corner detection algorithm by running the algorithm on publicly available datasets of \cite{Dataset}, \cite{dataset2}. We carefully selected a number of subsets to ensure a fair and comprehensive evaluation scheme. The selected subsets composed of simple and complex scenes, including low and high textured environments, and slow and fast motions. The event camera generates few events per second when the camera and objects are moving slowly. On the other hand, in highly dynamic and complex scenes, the number of events per second can reach up to 10 million. The datasets are recorded by a DAVIS-240C \cite{davis}, which contain many sequences of frame-based, i.e., intensity images and asynchronous events at the resolution of 240x180. Note that the intensity images are only used to obtain the ground-truth for evaluation purposes.  The proposed algorithm has been implemented in software in C language; the controller implemented in~\cite{dac2018} has been integrated into the software to measure the throughput. Moreover, the message size was set to 10,000 events. The application was run on an embedded system with an ARM Cortex-A57 CPU at 2GHz clock frequency.

\subsection{Accuracy}

We adopted a similar technique proposed in \cite{arc} to evaluate our method against different competitive methods. The accuracy is calculated by TEC/(TEC+FEC), where True Event Corners (TEC) are event-corners that fall inside the small (inner) cylinder with a 3-pixel radius,  and False Event Corners (FEC) are event-corners that fall between the small cylinder and the large cylinder (5 pixels radius). The two oblique cylinders are constructed by computing the intensity-corners using Harris \cite{harris} and then track corners over the captured images using Kanade–Lucas–Tomasi KLT \cite{klt} to obtain the tracking line for each corner. This line is then used as the center of the two oblique cylinders.

Table \ref{tab:acc} shows that our method favorably compared to eFAST, Arc*, and FA-Harris. eFast and Arc* methods have the lowest accuracy since they only perform a pixel-wise comparison to detect corners.

\begin{table}[h]
\centering
\begin{threeparttable}
\caption{The accuracy [\%] of different methods. }\label{tab:acc}
\begin{tabular}{l|cccccc}
\hline
                     \textbf{Method} &\textit{shapes} &\textit{boxes} &\textit{walking} &\textit{run} &\textit{office}  
                    \\ \hline
\textbf{eFAST}\cite{eFAST}    &   56.40 & 48.59 & 51.09 & 55.9 & 54.59 \\                  
\textbf{eHarris}\cite{eharris}     & 57.01  &49.26 &69.26 &62.26  &61.3    \\
\textbf{Arc*}  \cite{arc} &   55.38 & 49.01 & 52.41 & 53.41 & 51.21 \\ 
 \textbf{FA-Harris}   \cite{fa-harris}&  57.66 & 49.66 & 65.32 & 49.66 & 63.66\\ 
\textbf{Ours}   &\textbf{63.20} & \textbf{53.27} & \textbf{72.1} & \textbf{68.7} & \textbf{69.62}\\   \hline

\end{tabular}
\end{threeparttable}
\end{table}

\subsection{Qualitative Evaluation}

\begin{figure}[ht]
\centering
\subfigure[raw]{\includegraphics[width = 1in]{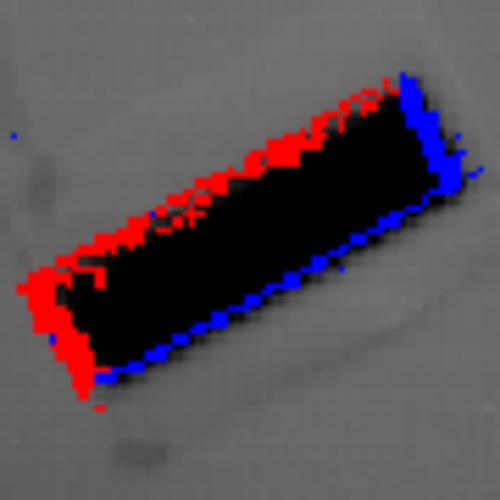}} \hspace{4pt}
\subfigure[Harris]{\includegraphics[width = 1in]{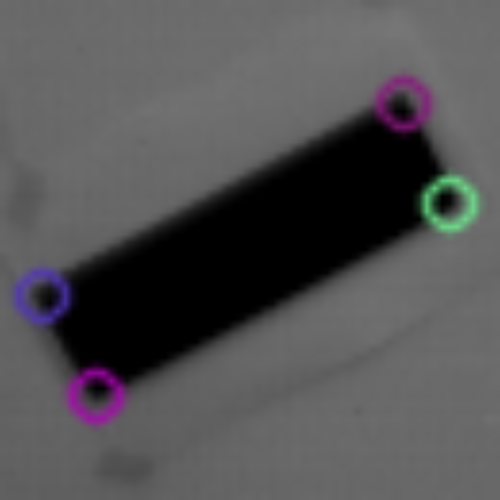}} \hspace{4pt}
\subfigure[eFAST]{\includegraphics[width = 1in]{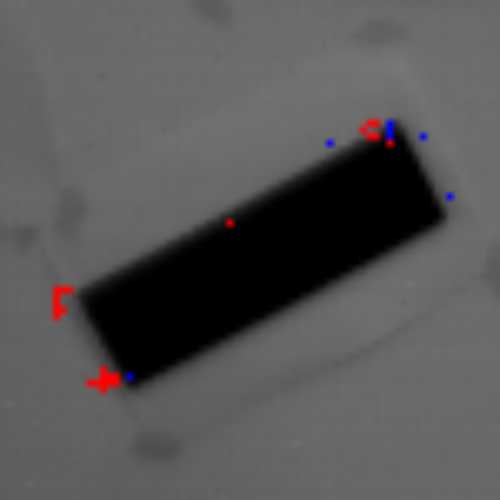}}\\ 
\subfigure[Arc*]{\includegraphics[width = 1in]{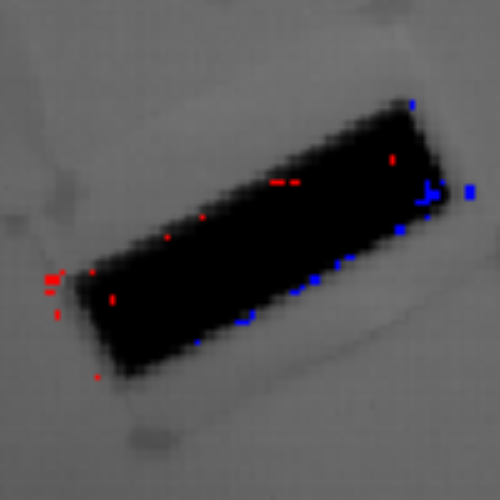}} \hspace{4pt}
\subfigure[eHarris]{\includegraphics[width = 1in]{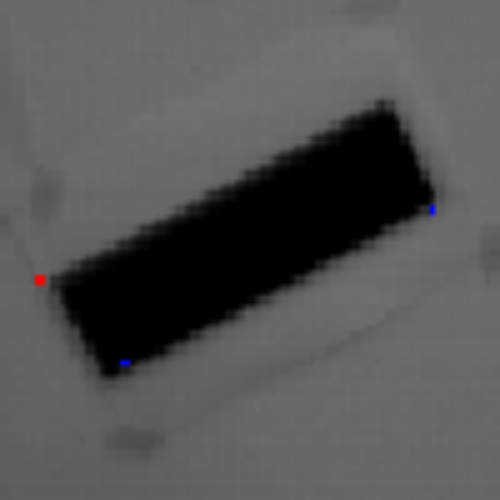}}\hspace{4pt}
\subfigure[Ours]{\includegraphics[width = 1in]{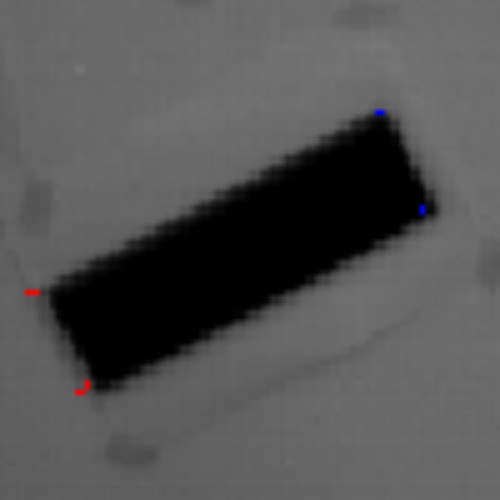}}
\caption{Qualitative comparison against different methods on the event camera dataset ($shapes\_6dof$). Our method performs comparably to eHarris \cite{eharris}, and produce high quality corners than eFAST\cite{eFAST} and Arc* \cite{arc} }
\label{fig:observe}
\end{figure}

Figure \ref{fig:observe} shows a qualitative comparison of different methods in the state-of-the-art. It shows that our method outperforms other algorithms in terms of the quality of corners and the number of event-corners per intensity-corner. In Figure \ref{fig:observe}.c, most of the detected event-corners by the eFAST are related to intensity-corners, however, the number of event-corners per intensity-corners is high. The main issue of the Arc* algorithm is that it mistakenly detects many false event-corners. The eHarris method (see Figure \ref{fig:observe}.e) also shows high-quality detection, however, it failed to detect the upper-right corner of the rectangle. 

In Table \ref{tab:filter}, we reported the reduction rate of our algorithm against different state-of-the-art methods. The reduction rate is the number of detected event-corners divided by the total amount of events and it defines the ability of a method to discard unnecessary events from the system pipeline. The results show that our algorithm is able to highly reduce the number of event-corners and only detect high-quality event-corners, thanks to the three-layer filtering and the Harris-based selector.

\subsection{Throughput Performance}
\begin{figure}[ht]
    \centering
    \includegraphics[width=1\columnwidth]{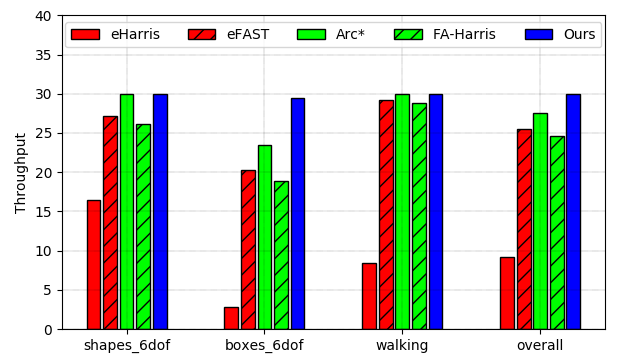}
    \caption{The throughput performance of different methods.}
    \label{fig:thr}
\end{figure}

One of the most important aspects of an application that is targeted for embedded systems is real-time performance, i.e., achieving the desired throughput under various conditions. The report of the average throughput for different algorithms is summarized in Figure \ref{fig:thr}. Our method is able to achieve the desired throughput (throughput = 30) in different scenarios. Other methods such as  Arc$^{*}$, eFAST, and FA$\_$Harris are only able to achieve close to the desired throughput in low-textured and slow-motion speed scenarios (e.g., shapes and walking). However, they failed to achieve the desired throughput in a high-textured environment (e.g., boxes$\_$6dof). The eHarris method fails to achieve the desired throughput in all case scenarios since it processes all incoming events and performs expensive operations.

{}

\subsection{LC\_Harris performance}

Table \ref{tab:exe} summarizes the execution time of our proposed LC\_Harris compared with eHarris \cite{eharris}. We evaluated both algorithms on five different datasets which includes fast and slow camera motion, static and dynamic environment, low-textured and textured environment and day and night scenes. Thanks to the approximation our algorithm is able to execute faster than the eHarris algorithm in all scenarios. LC\_Harris is able to achieve an average execution time savings of $59\%$ compared with eHarris.

\begin{table}[h]
\centering
\begin{threeparttable}
\caption{The average execution time [ns] of our algorithm LC\_Harris compared with Harris on different datasets. }\label{tab:exe}
\begin{tabular}{l|cccccc}
\hline
                     \textbf{Method} &\textit{shapes} &\textit{boxes} &\textit{walking} &\textit{run} &\textit{office} 
                    \\ \hline

\textbf{Harris}\cite{harris}     & 91 & 205 & 93 & 94 & 110  \\

\textbf{Ours}   &\textbf{56} & \textbf{110}& \textbf{57}& \textbf{55}& \textbf{66}\\ \hline

\end{tabular}
\end{threeparttable}
\end{table}

\section{Conclusions}
We have presented TLF-Harris, a fast and adaptive corner detection method that detects high-quality corners based on asynchronous events. We demonstrated that our approach facilitates execution of vision tasks on low-cost resource-constrained computing platforms that do not have sufficiently powerful processing units to handle millions of events per second on average. Our method is composed of a three-layer filtering stage and a low-complexity Harris score. In the first layer, we perform timestamp filtering, in which the filtering threshold is determined based on the throughput feedback and the optical flow of the event.  In the second layer, an arc filter is used to perform a pixel-wise comparison on the eSAE to keep only corner candidates. The third filter, i.e., the lifetime filter uses the C-SAE to measure the lifetime of the event and avoid processing any incoming corner candidates in the neighborhood until the old corner candidate fades, i.e., the timestamp of the incoming event is later than the lifetime of the old corner candidate. A low-complex Harris is proposed to compute the score and decide whether the incoming event is a corner or not. Our method is able to maintain the desired throughput in various scenarios such as low-textured, high-textured, and fast motion speed scenarios.
Moreover, our method performs favorably compared with state-of-the-art methods in terms of accuracy.

\bibliographystyle{IEEEtran}
\bibliography{ref}

\end{document}